\documentclass[journal,12pt,onecolumn]{IEEEtran}
\usepackage{amsmath}
\usepackage{amsfonts}
\usepackage{algorithm, algpseudocode}

\usepackage{microtype}
\usepackage{graphicx}
\usepackage{subfigure}
\usepackage{booktabs} 

\newcommand{\vx}{\mathbf{x}}  
\newcommand{\features}{\vx}  
\newcommand{\featurelen}{n} 
\newcommand{\eermparam}{\lambda}
\newcommand{\hypothesis}{h}  
\newcommand{\samplesize}{m}  
\newcommand{\dataset}{\mathcal{D}}  
\newcommand{\sampleidx}{i}
\newcommand{\defeq}{:=}
\newcommand{\truelabel}{y}  
\newcommand{\weights}{\mathbf{w}} 
\newcommand{\weight}{w} 
\newcommand{\user}{u} 
\newcommand{\feature}{x} 
\newcommand{\treedepth}{d} 
\newcommand{\featureidx}{j} 
\newcommand{\lagmult}{\rho} 
\newcommand{\risk}[1]{\overline{L}(#1)}
\newcommand{\hypospace}{\mathcal{H}} 
\newcommand{\emprisk}{\widehat{L}}
\newcommand{\loss}[2]{L\big({#1},{#2}\big)} 
\DeclareMathOperator*{\argmin}{arg\;min}
\newcommand{\expect}{{\rm E}}
\newcommand{\subjectiveexplainability}[2]{E(#1|#2)}
\usepackage{tikz}
\usepackage{pgfplots}
\usetikzlibrary{positioning,shapes,arrows}
\usetikzlibrary{arrows}
\usetikzlibrary{positioning}
\usetikzlibrary{shapes.geometric}
\usetikzlibrary{backgrounds}

\usepackage{hyperref}
\usepackage{xspace}

\usepackage{amsmath}
\usepackage{amssymb}
\usepackage{mathtools}
\usepackage{amsthm}

\usepackage[capitalize,noabbrev]{cleveref}

\theoremstyle{plain}

\theoremstyle{definition}

\theoremstyle{remark}

\usepackage[textsize=tiny]{todonotes}

\begin{document}

\title{Explainable Empirical Risk Minimization}
%
%
%

\author{L.~Zhang$^{1}$, G.~Karakasidis$^{1}$, A.~Odnoblyudova$^{1}$, L.~Dogruel$^{2}$ and A.~Jung$^{1}$
	\thanks{$^{1}$ Aalto University, Espoo, Finland}
	\thanks{$^{2}$ Johannes Gutenberg-Universität Mainz, Mainz, Germany}
}

\maketitle

\markboth{Some Journal}%
{Shell \MakeLowercase{\textit{et al.}}: Bare Demo of IEEEtran.cls for IEEE Journals}

%




\begin{abstract}
The successful application of machine learning (ML) methods becomes increasingly dependent  
on their interpretability or explainability. Designing explainable ML systems is instrumental to 
ensuring transparency of automated decision-making that targets humans. The explainability of 
ML methods is also an essential ingredient for trustworthy artificial intelligence. 
A key challenge in ensuring explainability is its dependence on the specific human user (``explainee''). 
The users of machine learning methods might have vastly different background knowledge 
about machine learning principles. One user might have a university degree in machine learning 
or related fields, while another user might have never received formal training in high-school 
mathematics. This paper applies information-theoretic concepts to develop a novel measure for 
the subjective explainability of the predictions delivered by a ML method. We construct this measure 
via the conditional entropy of predictions, given a user feedback. The user feedback might be obtained from 
user surveys or biophysical measurements. Our main contribution is the explainable empirical risk 
minimization (EERM) principle of learning a hypothesis that optimally balances between the subjective 
explainability and risk. The EERM principle is flexible and can be combined with arbitrary machine 
learning models. We present several practical implementations of EERM for linear models and 
decision trees. Numerical experiments demonstrate the application of EERM to detecting the use of 
inappropriate language on social media. 
\end{abstract}

\section{Introduction}
\label{sec_intro}

We consider machine learning (ML) methods that learn a hypothesis map that reads in features of a data point 
and outputs a prediction for some quantity of interest (label). Explainable ML (XML) aims at supporting human 
end users in understanding (at least partially) how a ML method arrives at its final predictions \cite{Ribeiro2016,Holzinger2018,Hagras2018,Mittelstadt2016,Wachter2017,Chen2018,Roscher2020}. 
One key challenge of XML is the variation in background knowledge of human end users \cite{BelleXML2021,JunXML2020}. 
ML methods that are explainable for a domain expert might be opaque (``black-box'') for a lay user. 

As a point in case, consider a ML method that uses a deep net to diagnose skin cancer from images \cite{Esteva2017}. 
The predictions obtained from a deep net which is fed by an image, might be explained by quantifying 
the influence of individual pixels on the resulting prediction \cite{Ayhan2022}. We can conveniently visualize 
such influence measures as a saliency map \cite{Bach2015}. While the so-obtained saliency map is a useful explanation 
to a dermatologist, it might not offer a sufficient explanation for a lay person without a university-level training in dermatology. 

There seems to be no widely accepted definition for the (level of) explainability of a ML method \cite{Linardatos2021,Zhou2021}. 
Basic requirements for explainable ML include stability (obtaining similar predictions 
in similar cases) and interpretability (relating predictions to known concepts) $\ldots$ \cite{Burkart2021,Roscher2020}. 
We try to capture important aspects of explainable ML by identifying explainability with a notion of 
predictability or the lack of uncertainty \cite{Roscher2020}. In particular, we use a subjective (or personalized) 
notion of explainability that reflects the discrepancy between ML predictions and the intuition 
of a specific user. We encode this intuition by a user-specific notion of similarity between data points. 
Subjective explainability of ML method requires similar predictions for data points considered similar by a user. 


This paper proposes explainable empirical risk minimization (EERM) as a novel XML method. The main 
idea behind EERM is to learn a hypothesis whose predictions do not deviate too much across data points 
that are considered similar by the human consumer of these predictions (``end user''). 
To this end, we require the user to provide a feedback signal for data points in a training set. 
We allow for wide range of possible feedback signals. The user feedback might be answers to a survey, 
bio-physical measurements or visual observations of facial expressions \cite{ZUBAREV2022100951,Ranga:2020uw,Strandvall2009}. 
The goal is to learn a hypothesis that conforms with the user intuition in the sense of delivering similar 
predictions for data points with similar user feedback signals. Our approach is somewhat related to 
prototype or example-based explanations with candidate prototypes (or examples) being data points 
having similar user feedback signals \cite{NIPS2016_5680522b}. 

It is important to note that our approach allows for arbitrary user feedback signals. We might call 
our method as ``user-agnostic'' as we only require the user feedback signal for data points in a training set. 
Besides the training set we do not require any input from or information about the user. Depending on 
the quality of the user feedback signal, enforcing subjective explainability of a learnt hypothesis might be
beneficial or detrimental to the resulting prediction accuracy (see Section \ref{sec_explainable_linreg}). 
If the user feedback is strongly correlated with the true label of a data point, our requirement for 
explainability helps to steer or regularize the learning task. Indeed, we might interpret the user 
feedback signal as a manifestation of domain expertise and, in tun, the explainability requirement 
as a means to incorporate domain expertise into a ML method.


\subsection{State of the Art} 
The increasing need for explainable ML sparked a considerable amount of research on different methods for 
explainable ML \cite{Linardatos2021}. Methods for explainable ML can be categorized based on 
different aspects \cite{Linardatos2021,Zhou2021}. One important aspect is the restrictions placed 
on the underlying ML models. In this regard, there are two orthogonal approaches to 
explainable ML: either use  an intrinsically-explainable (``simple``) model or model-agnostic methods 
that explicitly generate explanations for a given (``black-box'') ML method \cite{Molnar2019}. 
Model-agnostic methods provide post-hoc (after model training) explanations \cite{Ribeiro2016,JunXML2020}. 
These methods do not require the details of a ML method but only require its predictions for some training examples. 

Examples of intrinsically explainable models include linear models using few features and shallow decision 
trees \cite{Molnar2019}. The interpretation of a linear model is typically obtained from an inspection of the 
learned weights for the individual features. A large (in magnitude) weight is then read as an indicator 
for a high relevance of the corresponding feature. The prediction delivered by a decision tree might be in 
the form of the path from root node to the decision node (essentially a sequence elementary tests on the 
features of the data point). In general, however, there is no widely accepted definition of which model is 
considered intrinsically explainable. Moreover, there is no consensus about how to measure the 
explainability of a ``simple'' model \cite{Zhou2021}. 

A main contribution of this paper is a method for constructing an explainable model by regularizing a 
given (high-dimensional) ML model \cite{Montavon2018,Bach2015,Hagras2018}. As regularization term, 
we use a novel measure for subjective explainability of predictions obtained from a hypothesis map. 
What sets this work apart from most existing work on explainable ML is that we use a novel measure of subjective 
explainability. This measure is implemented using the concept of a user feedback. Broadly speaking, the 
user feedback is some user-specific attribute that is assigned or associated with a data point. Formally, we 
can think of the user feedback signal as an additional (user-specific) feature of a data point. This additional 
feature is measured or determined via the user and revealed to our method for each data point. 

Similar to \cite{Chen2018}, we use information-theoretic concepts to measure subjective explainability. 
However, while \cite{Chen2018} uses the mutual information between an explanation and the prediction, 
we measure the subjective explainability of a hypothesis using the conditional entropy of its predictions 
given a user feedback signal. This conditional entropy is then used as a regularizer for empirical risk 
minimization (ERM) resulting in explainable empirical risk minimization (EERM). 

The EERM principle requires a training set consisting of data points for which, beside their features, 
also the label and user signal values are known. The user signal values for the data points in the training 
set are used to estimate the subjective explainability of a hypothesis. We obtain different instances of EERM 
form different hypothesis spaces (models). Two specific instances are explainable linear regression 
(see Section \ref{sec_explainable_linreg}) and explainable decision tree classification (see Section \ref{sec_dectree}). 

We illustrate the usefulness of EERM using the task of detecting hate speech in social media. 
Hate speech is a main obstacle towards embracing the Internet’s potential for deliberation and 
freedom of speech \cite{Laaksonen2020}. Moreover, the detrimental effect of hate speech 
seems to have been amplified during the current Covid-19 pandemic \cite{Hardage2020}. 
Detecting hate speech requires multi-disciplinary expertise from both social science and computer 
science expertise \cite{Papcunova:2021vf,LiaoXAI2021}. Providing subjective explainability for ML users 
with different backgrounds is crucial for the diagnosis and improvement of hate speech 
detection systems \cite{Laaksonen2020,Hardage2020,Bunde2021}.

\subsection{Contributions} 
Our main contributions can be summarized as follows: 
\begin{itemize} 
	\item We introduce a novel measure for the subjective explainability of the predictions delivered by a ML 
	          method to a specific user. This measure is constructed from the conditional entropy of the 
	          predictions given some user signal (see Section \ref{sec_explainability}). 
	
	\item Our main methodological contribution is EERM which uses subjective explainability as regularizer. We 
	          present two equivalent (dual) formulations of EERM as optimization problems (see Section \ref{sec_eerm}). 
	          
	  \item We detail practical implementations of the EERM principle for linear regression and decision 
	            tree classification (see Section \ref{sec_explainable_linreg} - \ref{sec_dectree}).
	  
	  \item We illustrate the usefulness of the EERM principle by some illustrative numeric experiments. 
	  These experiments revolve around explainable weather forecasting and explainable hate-speech 
	  detection in social media (see Section \ref{sec_num_exp}). We use EERM to learn 
	  an explainable decision tree classifier for a user that associates hate speech with the 
	  presence of specific keywords.  
\end{itemize}  


\section{Problem Setup}
\label{sec_setup}

We consider a ML application that involve data points, each characterized by a label (quantity of interest) 
$\truelabel$ and some features (attributes) $\features = \big(\feature_{1},\ldots,\feature_{\featurelen} \big)^{T} \in \mathbb{R}^{\featurelen}$ \cite{hastie01statisticallearning,BishopBook}. ML methods aim at learning a hypothesis map $h$ that allows 
to predict the label of a data point based solely on its features. 

In contrast to standard ML approaches, we explicitly take the specific user of the ML method into account. 
Each data point is also assigned a user signal $\user$ that characterizes it from the perspective of a 
specific human user. The concept of a user signal $\user$ is similar to the features $\features$ of a data point. 
Like features, also a user signal is a quantity or property of a data point that can be measured easily in an 
automated fashion. However, while features typically represent objective measurements of physical quantities, 
the user signal $\user$ is a subjective measurement provided (actively or passively) by the human user 
of the ML method. 

Let us illustrate the rather abstract notion of a user signal by some examples. One important 
example for a user signal is a manually constructed feature of the data point. Section \ref{sec_num_exp} 
considers hate speech detection in social media where data points represent short messages (``tweets''). 
Here, the user signal $\user$ for a specific data point could be defined via the presence of a 
certain word that is considered a strong indicator for hate speech.

The user signal $\user$ might also be collected in a more indirect fashion. Consider 
an application where data points are images that have to be classified into different categories. 
Here, the user signal $\user$ might be derived from EEG measurements taking when a data point 
(image) is revealed to the user \cite{ZUBAREV2022100951}.

The goal of supervised ML is to learn a hypothesis 
\begin{equation} 
	\label{equ_pred_map}
	h(\cdot): \mathbb{R}^{\featurelen} \rightarrow \mathbb{R}: \features \mapsto \hat{\truelabel}=\hypothesis(\features). 
\end{equation}  
that is used to compute the predicted label $\hat{\truelabel}=\hypothesis(\features)$ from the 
features $\features = \big(\feature_{1},\ldots,\feature_{\featurelen}\big)^{T} \in \mathbb{R}^{\featurelen}$ of a data point. 
Any ML method that can only use finite computational resources can only use  a subset of (computationally) 
feasible maps. We refer to this subset as the hypothesis space (model) $\mathcal{H}$ of a ML method. 
Examples for such a hypothesis space are linear maps, decision trees or artificial neural networks \cite{HastieWainwrightBook,Goodfellow-et-al-2016}. 

For a given data point with features $\features$ and label $\truelabel$, we measure the quality of a 
hypothesis $h$ using some loss function $L((\features,\truelabel),\hypothesis)$. 
The number $\loss{\big(\features,\truelabel\big)}{h}$ measures the error incurred by 
predicting the label $\truelabel$ of a data point using the prediction $\hat{\truelabel}= h(\features)$. 
Popular examples for loss functions are the squared error loss $\loss{\big(\features,\truelabel\big)}{h} = (h(\features) - \truelabel)^{2}$ 
(for numeric labels $\truelabel \in \mathbb{R}$) or the logistic loss $\loss{\big(\features,\truelabel\big)}{h}  = \log(1+\exp(-h(\features)\truelabel))$ 
(for binary labels $\truelabel \in \{-1,1\}$).

Roughly speaking, we would like to learn a hypothesis $h$ that incurs small loss on any data point. 
To make this informal goal precise we can use the notion of expected loss or risk 
\begin{equation}
\label{equ_def_risk}
\risk{\hypothesis} \defeq \expect \big\{  \loss{\big(\features,\truelabel\big)}{\hypothesis}  \big\}. 
\end{equation} 
Ideally, we would like to learn a hypothesis $\hat{\hypothesis}$ with minimum risk 
\begin{equation} 
\label{equ_def_risk_min}
\risk{\hat{\hypothesis}} = \min_{h \in \hypospace} \risk{\hypothesis}. 
\end{equation}
It seems natural to learn a hypothesis by solving the risk minimization problem \eqref{equ_def_risk_min}. 

There are two caveats to consider when using the risk minimization principle \eqref{equ_def_risk_min}. 
First, we typically do not know the underlying probability distribution $p(\features,\truelabel)$ required for 
evaluating the risk \eqref{equ_def_risk}. We will see in Section \ref{sec_def_erm} how empirical risk 
minimization (ERM) is obtained by approximating the risk using an average loss over some training set. 

The second caveat to a direct implementation of risk minimization \eqref{equ_def_risk_min} is 
its ignorance about the explainability of the learned hypothesis $\hat{h}$. 
In particular, we are concerned with the subjective explainability of the predictions $\hat{h}(\features)$ 
for a user that is characterized via a user signal $\user$ for each data point. We construct a 
measure for this subjective explainability in Section \ref{sec_explainability} and use it as a 
regularizer to obtain explainable ERM (EERM) (see Section \ref{sec_eerm}).

\subsection{Empirical Risk Minimization}
\label{sec_def_erm}

The idea of ERM is to approximate the risk \eqref{equ_def_risk} using the average loss (or empirical risk) 
\begin{equation} 
	\label{equ_def_ER}
\emprisk (h| \dataset) \defeq (1/\samplesize) \sum_{\sampleidx=1}^{\samplesize}  \loss{\big(\features^{(\sampleidx)},\truelabel^{(\sampleidx)} \big)}{h}. 
\end{equation}
The average loss $\emprisk (h| \dataset)$ of the hypothesis $h$ is measured on a set of labelled 
data points (the training set)
\begin{equation}
	\label{equ_def_dataset}
	\dataset= \big\{ \big(\features^{(1)},\truelabel^{(1)}, \user^{(1)}\big),\ldots,\big(\features^{(\samplesize)},\truelabel^{(\samplesize)}, \user^{(\samplesize)} \big) \big\}. 
\end{equation}
The training set $\dataset$ contains data points for which we know the true label value $\truelabel^{(\sampleidx)}$ 
and the corresponding user signal $\user^{(\sampleidx)}$. 

Section \ref{sec_num_exp} applies our methods to the problem of hate speech detection. In this application, 
a data point is a short text message (``tweet'') and the training set \eqref{equ_def_dataset} consists of 
tweets for which we know if they are hate speech or not. As the user signal we will use the presence of a 
small number of keywords that are considered a strong indicator for hate speech. 

Many practical ML methods are based on solving the ERM problem 
\begin{equation} 
	\label{equ_def_ERM}
	\hat{h} \in \argmin_{h \in \hypospace}\emprisk (h| \dataset) . 
\end{equation} 
However, a direct implementation of ERM \eqref{equ_def_ERM} is prone to overfitting if the hypothesis space $\hypospace$ 
is too large (e.g., linear maps using many features and very deep decision trees) compared to the size $\samplesize$ 
of the training set. To avoid overfitting in this high-dimensional regime \cite{BuhlGeerBook,Wain2019}, 
we add a regularization term $ \eermparam \mathcal{R}(h)$ to the empirical risk in \eqref{equ_def_ERM},
\begin{equation}
	\label{equ_def_rerm}
	h^{(\eermparam)} \in \argmin_{h \in \hypospace} \emprisk(h| \dataset)+ \eermparam \mathcal{R}(h). 
\end{equation}  
The choice of the regularization parameter $\eermparam\!\geq\!0$ in \eqref{equ_def_rerm} can be guided by a 
probabilistic model for the data or using validation techniques \cite{hastie01statisticallearning}. 

A dual form of regularized ERM \eqref{equ_def_rerm} is obtained by replacing the regularization 
term with a constraint, 
\begin{equation}
	\label{equ_def_rerm_constrained}
	h^{(\eta)} \in \argmin_{h \in \hypospace} \emprisk(h| \dataset) \mbox{ such that }   \mathcal{R}(h)\leq \eta. 
\end{equation}  
The solutions of \eqref{equ_def_rerm_constrained} coincide with those of \eqref{equ_def_rerm} for an appropriate 
choice of $\eta$ \cite{BertsekasNonLinProgr}. Solving the primal formulation \eqref{equ_def_rerm} might be 
computationally more convenient as it is an unconstrained optimization problem in contrast to the dual formulation \eqref{equ_def_rerm_constrained} \cite{BoydConvexBook}. However, the dual form \eqref{equ_def_rerm_constrained} allows to explicitly specify 
an upper bound $\eta$ on the value $\mathcal{R}(h^{(\eta)})$ for the learned hypothesis $h^{(\eta)}$. 


Regularization techniques are typically used to improve statistical performance (risk) of the learned hypothesis. 
Instead, we use regularization as a vehicle for ensuring explainability. In particular, we do not use the regularization 
term as an estimate for the generalization error $\risk{\hypothesis}- \emprisk(h| \dataset)$. Rather, we use a 
regularization term that measures for the subjective explainability of the predictions $\hat{\truelabel}=h(\features)$. 
The regularization parameter $\eermparam$ in \eqref{equ_def_rerm} (or $\eta$ in the dual formulation \eqref{equ_def_rerm_constrained}) 
adjusts the level of subjective explainability of the learned hypothesis $\hat{\hypothesis}$. 



\subsection{Subjective Explainability}
\label{sec_explainability}

There seems to be no widely accepted formal definition for the explainability (interpretability) of 
a learned hypothesis $\hat{h}$. Some authors simply define specific ML methods to deliver a 
explainable hypothesis \cite{Molnar2019}. Examples for such intrinsically explainable ML methods include 
linear regression and (shallow) decision trees. 

While linear regression is sometimes considered as interpretable, the predictions obtained by applying 
a linear hypothesis to a huge number of features might be difficult to grasp. Moreover, the interpretability 
of linear models also depends on the background (formal training) of the specific user of a ML method. 

Similar to \cite{Chen2018} we use information-theoretic concepts to make the notion of explainability 
precise. This approach interprets each data point as realizations of i.i.d. random variables. In particular, 
the features $\features$, label $\truelabel$ and user signal $\user$ associated with a data point 
are realizations drawn from a joint probability density function (pdf) $p(\features,\truelabel,\user)$. 
In general, the joint pdf $p(\features,\truelabel,\user)$ is unknown and needs to be estimated from data using, e.g., maximum 
likelihood methods \cite{BishopBook,HastieWainwrightBook}. 

\begin{figure}[htbp]
	\hspace*{0mm}
	\begin{center}
		\begin{tikzpicture}[
			node distance=1cm and 0cm,
			mynode/.style={draw,ellipse,text width=1.3cm,align=center}, 
			mynode1/.style={draw,ellipse,text width=1.6cm,align=center}
			]
			\node[mynode1] (dp) {data point $(\features,\truelabel,\user)$};
			\node[mynode,right=2cm of dp] (prediction) {prediction $\hat{\truelabel}$};
			
			\draw[->](dp)  -- (prediction)  node[midway, above] {$h \in \hypospace$};
		\end{tikzpicture}
	\end{center}
	\vspace*{0mm}
	\caption{The features $\features$, label $\truelabel$ and user signal $\user$ of a data point 
		are realizations drawn from a pdf $p(\features,\truelabel,\user)$. Our goal is to learn a hypothesis $h$ such 
		that its predictions $\hat{\truelabel}$ have a small conditional entropy given the user signal $\user$.}
	\label{fig_simple_prob_ML}
\end{figure}
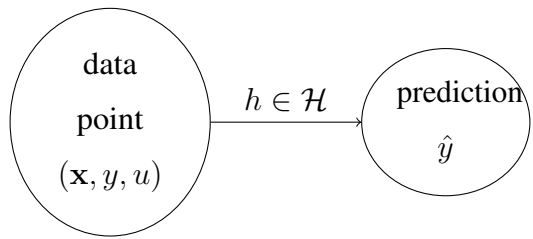

Note that since we model the features of a data point as the realization of a random variable, 
the prediction $\hat{\truelabel} = h(\features)$ also becomes the realization of a random variable. 
Figure \ref{fig_simple_prob_ML} summarizes the overall probabilistic model for data points, the 
user signal and the predictions delivered by (the hypothesis learned with) a ML method. 

We measure for the subjective explainability of the predictions $\hat{\truelabel}$ 
delivered by a hypothesis $h$ for some data point $\big(\features,\truelabel,\user\big)$ as, 
\begin{equation} 
	\label{eq_def_explainability}
	 \subjectiveexplainability{\hypothesis}{\user} \defeq C- H(\hypothesis|\user).
\end{equation} 
Here, we used the conditional (differential) entropy $H( h |\user)$ (see Ch.\ 2 and Ch.\ 8 \cite{coverthomas})
\begin{align} 
	H(h|\user) & \defeq - \expect\bigg\{ \log p(\underbrace{ h(\features) }_{=\hat{\truelabel}}|\user) \bigg\} 
\end{align} 
We introduce the (``calibration'') constant $C$ in \eqref{eq_def_explainability} for notational convenience. 
The actual value of $C$ is meaningless for our approach (see Section \ref{sec_eerm}) and serves only 
the convention that the subjective explainability $\subjectiveexplainability{\hypothesis}{\user}$ is non-negative. 

For regression problems, the predicted label $\hat{\truelabel}$ might be modelled as a continuous random 
variable. In this case, the quantity $H(\hat{\truelabel}|\user)$ is a conditional differential entropy. 
With slight abuse of notation we refer to $H(\hat{\truelabel}|\user)$ as a conditional entropy and do not explicitly 
distinguish between the case where $\hat{\truelabel}$ is discrete, such as in classification problems studied in 
Sections \ref{sec_explainable_linreg}-\ref{sec_dectree} and Section \ref{sec_num_exp}. 

The conditional entropy $H(\hypothesis|\user)$ in \eqref{eq_def_explainability} quantifies the 
uncertainty (of a user that assigns the value $\user$ to a data point) about the prediction $\hat{\truelabel} = h(\features)$ 
delivered by the hypothesis $\hypothesis$. Smaller values $H(h|\user)$ correspond to smaller levels of 
subjective uncertainty about the predictions $\hat{\truelabel} = \hypothesis(\features)$ 
for a data point with known user signal $\user$. This, in turn, corresponds to a larger value $ \subjectiveexplainability{\hypothesis}{\user}$ 
of subjective explainability. 

Section \ref{sec_num_exp} discusses explainable methods for detecting hate speech or the use of 
offensive language. A data point represents a short text message (a tweet). Here, the user signal $\user$ 
could be the presence of specific keywords that are considered a strong indicator for hate speech or 
offensive language. These keywords might be provided by the user via answering a survey or they 
might be determined by computing word histograms on public datasets that have been 
manually labeled \cite{Davidson2017}.


\section{Explainable Empirical Risk Minimization}
\label{sec_eerm}

Section \ref{sec_setup} has introduced all the components of EERM as a novel 
principle for explainable ML. EERM learns a hypothesis $h$ by using an estimate 
$\widehat{H}(\hypothesis|\user)$ for the conditional entropy in \eqref{eq_def_explainability} 
as the regularization term $\mathcal{R}(h)$ in \eqref{equ_def_rerm}, 
\begin{equation} 
\label{equ_def_eerm} 
\hypothesis^{(\eermparam)} \!\defeq\! \argmin_{h \in \hypospace} \emprisk(\hypothesis| \dataset)+ \eermparam \underbrace{ \widehat{H}(\hypothesis|\user)}_{= \mathcal{R}(h)}. 
\end{equation} 
A dual form of \eqref{equ_def_eerm} is obtained by specializing \eqref{equ_def_rerm_constrained}, 
\begin{equation} 
	\label{equ_def_eerm_constrained} 
\hypothesis^{(\eta)} \!\defeq\! 	\argmin_{\hypothesis \in \hypospace} \emprisk( \hypothesis| \dataset)  \mbox{ such that }   \widehat{H}(\hypothesis|\user) \leq \eta. 
\end{equation}
The empirical risk $\emprisk(\hypothesis| \dataset)$ and the regularizer $\widehat{H}(\hypothesis|\user)$ are computed 
solely from the available training set \eqref{equ_def_dataset}. We will discuss specific choices 
for the estimator $\widehat{H}(\hat{\truelabel}|\user)$ in Section \ref{sec_explainable_linreg} - \ref{sec_dectree}. 

The idea of EERM is that the solution of \eqref{equ_def_eerm} (or \eqref{equ_def_eerm_constrained}) is 
a hypothesis that balances the requirement of a small loss (accuracy) with a sufficient level of subjective 
explainability $\subjectiveexplainability{\hypothesis}{\user} \big(= C -  H(\hypothesis|\user))$. This balance 
is steered by the parameter $\eermparam$ in \eqref{equ_def_eerm} and $\eta$ in \eqref{equ_def_eerm_constrained}, 
respectively. 

\begin{figure}[h]
	\begin{center}
		\includegraphics[width=\columnwidth]{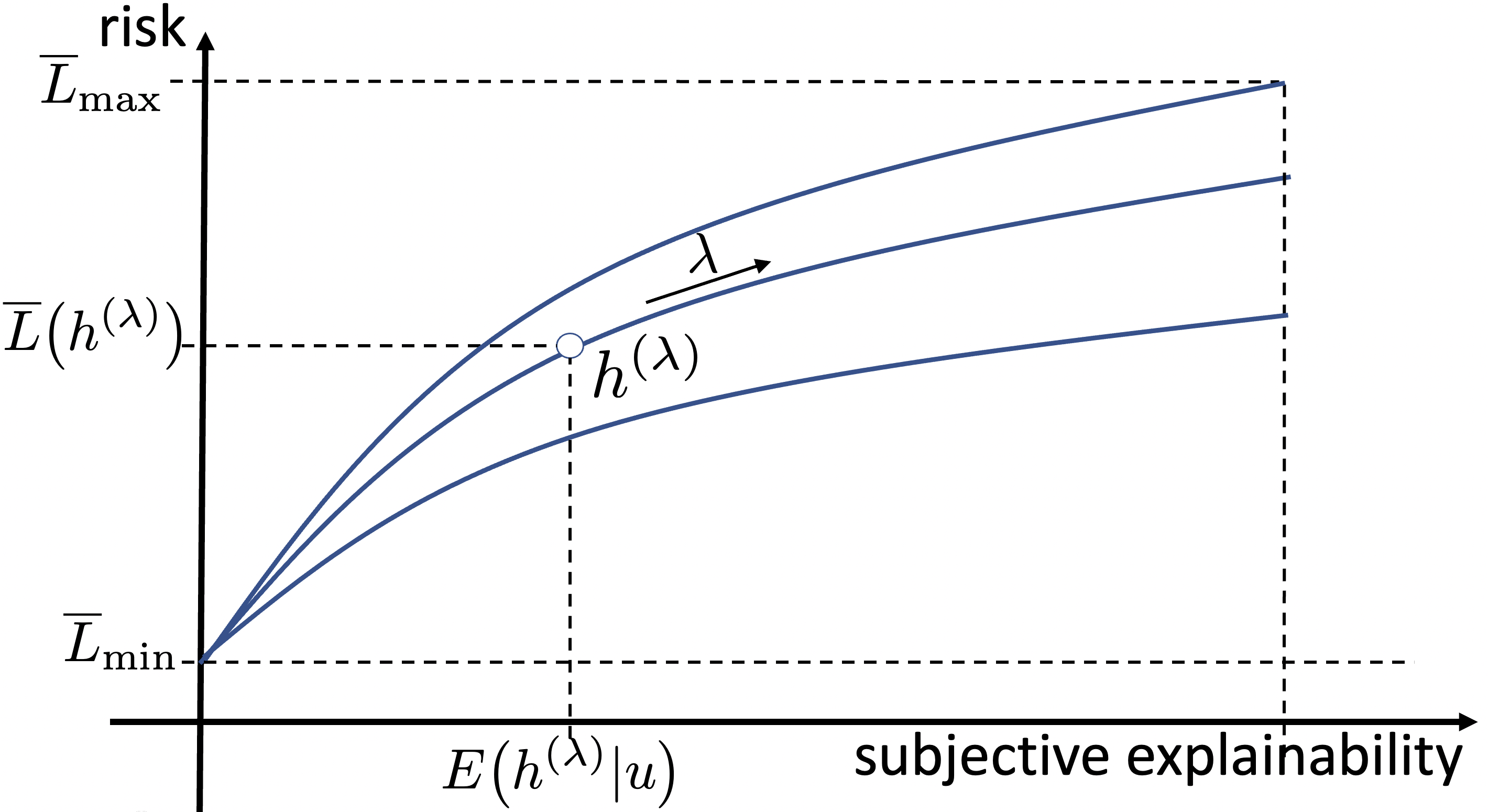}
		\vspace*{-3mm}
		\caption{The solutions of EERM, either in the primal \eqref{equ_def_eerm} or dual \eqref{equ_def_eerm_constrained} form, 
			trace out a curve in the plane spanned by the risk $\overline{L}(h)$ and the subjective explainability $E(h|\user)$.}
		\label{fig:eermcurve}
		\vspace*{-1mm}
	\end{center}
\end{figure}

Figure \ref{fig:eermcurve} illustrates the parametrized solutions of \eqref{equ_def_eerm} in the plane spanned 
by risk and subjective explainability. The different curves in Figure \ref{fig:eermcurve} are parametrized solutions 
of \eqref{equ_def_eerm} obtained from using different training sets (assumed to consists of i.i.d. data points) 
and different estimators $\widehat{H}$ of the conditional entropy $H(h|\user)$ in \eqref{eq_def_explainability}. 

Choosing a large value for $\eermparam$ in \eqref{equ_def_eerm} (small value for $\eta$ in \eqref{equ_def_eerm_constrained}) 
penalizes any hypothesis resulting in a large estimate $\widehat{H}(\hypothesis|\user)$ for the 
conditional entropy $H(\hypothesis|\user)$. Assuming $\widehat{H}(\hypothesis|\user) \approx H(\hypothesis|\user)$, 
using a large $\eermparam$ in \eqref{equ_def_eerm} (small $\eta$ in \eqref{equ_def_eerm_constrained} 
enforces a high subjective explainability \eqref{eq_def_explainability} of the learned hypothesis $\hypothesis^{(\eermparam)}$. 
Asymptotically (for $\eermparam \rightarrow \infty$), the solutions $\hypothesis^{(\eermparam)}$ 
of \eqref{equ_def_eerm} will maximize subjective explainability $E(\hypothesis|\user)$ at the cost of  
increasing risk. Figure \ref{fig:eermcurve} indicates the limit $\lim_{\lambda \rightarrow \infty} \overline{L}(\hypothesis^{(\eermparam)})$ 
as $\overline{L}_{\rm max}$. 




For the specific  choice $\eermparam =0$, EERM \eqref{equ_def_eerm} reduces to plain ERM that delivers a 
hypothesis $\hypothesis^{(\eermparam=0)}$ with risk $\overline{L}_{\rm min}$. This special case of EERM is obtained 
from the dual form \eqref{equ_def_rerm_constrained} using a sufficiently large $\eta$. The small risk of 
$\hypothesis^{(\eermparam=0)}$ comes at the cost of a relatively small subjective explainability $E(\hypothesis^{(\eermparam=0)}|\user)$. 
We choose the constant $C$ in \eqref{eq_def_explainability} such that $E(\hypothesis^{(\eermparam=0)}|\user)=0$ for notational 
convenience.

\subsection{Explainable Linear Regression} 
\label{sec_explainable_linreg}

We now specialize EERM in its primal form \eqref{equ_def_eerm} to linear regression \cite{BishopBook,HastieWainwrightBook}. 
Linear regression methods learn the parameters $\weights$ of a linear hypothesis $h^{(\weights)}(\features) = \weights^{T} \features$ to 
minimize the squared error loss of the resulting prediction error. The features $\features$ 
and user signal $\user$ of a data point are modelled realizations of jointly Gaussian random 
variables with mean zero and covariance matrix $\mathbf{C}$,
\begin{equation} 
	\label{equ_Gaussian_feature_summary}
	\big(\features^{T},\user\big)^{T} \sim \mathcal{N}(\mathbf{0},\mathbf{C}). 
\end{equation}  
Note that \eqref{equ_Gaussian_feature_summary} only specifies the marginal of the joint 
pdf $p(\features,\truelabel,\user)$ (see Figure \ref{fig_simple_prob_ML}). Using the probabilistic model 
\eqref{equ_Gaussian_feature_summary}, we obtain (see \cite{coverthomas})
\begin{align}
	\label{equ_sup_mi_Gauss}
	H(h|\user) &= (1/2) \log  \sigma^{2}_{\hat{\truelabel}|\user}. 
\end{align}
Here, we use the conditional variance $\sigma^{2}_{\hat{\truelabel}|\user}$ of $\hat{\truelabel} = h(\features)$ of 
the predicted label $\hat{\truelabel} = h(\features)$ given the user signal $\user$ for a data point.  

To develop an estimator $\widehat{H}(h|\user)$ for \eqref{equ_sup_mi_Gauss}, we use 
the identity \cite[Sec. 4.6.]{BertsekasProb}
\begin{equation} 
	\label{equ_def_optimal_coef_linmodel}
	 \sigma^{2}_{\hat{\truelabel}|\user}  = \min_{\alpha \in \mathbb{R}} \expect\big\{ \big(h(\features) - \alpha \user \big)^{2}\big\} . 
\end{equation} 
The identity \eqref{equ_def_optimal_coef_linmodel} relates the conditional variance $\sigma^{2}_{\hat{\truelabel}|\user}$ 
to the minimum mean squared error that can be achieved by estimating $\hat{\truelabel}$ using a 
linear estimator $\alpha \user$ with some $\alpha \in \mathbb{R}$. We obtain an estimator for the 
conditional variance $\sigma^{2}_{\hat{\truelabel}|\user}$ by replacing the expectation in 
\eqref{equ_def_optimal_coef_linmodel} by a sample average over the training set $\dataset$ \eqref{equ_def_dataset}, 
\begin{equation} 
	\label{equ_def_ent_estimator_linreg}
	\hat{\sigma}(\hat{\truelabel}|\user)  \defeq  \min_{\alpha \in \mathbb{R}} (1/\samplesize) \sum_{\sampleidx=1}^{\samplesize} \big( \weights^{T} \features^{(\sampleidx)} - \alpha \user^{(\sampleidx)}\big)^{2}.
\end{equation} 

It seems reasonable to estimate the conditional entropy $\widehat{H}(h^{(\weights)}|\user)$ via the 
plugging in the estimated conditional variance \eqref{equ_def_ent_estimator_linreg} into \eqref{equ_sup_mi_Gauss}, yielding 
the plug-in estimator $(1/2) $. However, in view of the duality between \eqref{equ_def_rerm_constrained} and \eqref{equ_def_eerm_constrained}, 
any monotonic increasing function of a given entropy estimator essentially amounts to a reparametrization 
$\lambda \mapsto \lambda'$ and $\eta \mapsto \eta'$. Since such a reparametrization is irrelevant as we choose 
$\lambda$ in a data-driven fashion, we will use the estimated conditional variance \eqref{equ_def_ent_estimator_linreg} 
itself as an estimator
\begin{equation} 
\label{equ_def_ent_estimator_linreg_estimator}
\widehat{H}(\hypothesis^{(\weights)}|\user)  \defeq  \min_{\alpha \in \mathbb{R}} (1/\samplesize) \sum_{\sampleidx=1}^{\samplesize} \big( \weights^{T} \features^{(\sampleidx)} - \alpha \user^{(\sampleidx)}\big)^{2}.
\end{equation} 
Note that we neither require the estimator \eqref{equ_def_ent_estimator_linreg_estimator} to be consistent 
nor to be unbiased \cite{LC}. Our main requirement is that, with high probability, the estimator \eqref{equ_def_ent_estimator_linreg_estimator} 
varies monotonically with the conditional entropy $H(\hypothesis^{(\weights)}|\user)$. 

Inserting the estimator \eqref{equ_def_ent_estimator_linreg_estimator} into EERM \eqref{equ_def_eerm}, yields 
Algorithm \ref{alg:explainable_linreg} as an instance of EERM for linear regression. Algorithm \ref{alg:explainable_linreg} 
requires as input a choice for the regularization parameter $\eermparam > 0$ and a training set $\dataset = \big\{ \big( \features^{(1)},\truelabel^{(1)} ,\user^{(1)} \big),\ldots,\big( \features^{(\samplesize)},\truelabel^{(\samplesize)} ,\user^{(\samplesize)} \big) \big\} $. 
As its output, Algorithm \ref{alg:explainable_linreg} delivers a hypothesis $h^{(\eermparam)}$ that compromises 
between small risk $\risk{\hypothesis}$ and subjective explainability $\subjectiveexplainability{\hypothesis}{\user}$. This 
compromise is controlled by the value of $\eermparam$. 

Choosing a large $\eermparam$ for Algorithm \ref{alg:explainable_linreg} favours a hypothesis $\hypothesis^{(\eermparam)}$ 
with small conditional entropy $H(\hypothesis^{(\eermparam)}|\user)$ and, in turn, high subjective explainability $\subjectiveexplainability{\hypothesis^{(\eermparam)}}{\user}$ (see \eqref{eq_def_explainability}). On the contrary, 
choosing a small $\eermparam$ puts more emphasis a small risk $\risk{\hypothesis^{(\eermparam)}}$ at the expense 
of increased conditional entropy $H(\hypothesis^{(\eermparam)}|\user)$ and, in turn, reduced subjective explainability $\subjectiveexplainability{\hypothesis^{(\eermparam)}}{\user}$.

\begin{algorithm}[htbp]
	\caption{Explainable Linear Regression}\label{alg:explainable_linreg}
	\begin{algorithmic}
		\State {\bfseries Input:} explainability parameter $\eermparam$, training set $\dataset$ (see \eqref{equ_def_dataset})
		\State 1: solve 
		\begin{align}
			\label{equ_P0}
			\hspace*{-7mm}\widehat{\weights} & \!\in\! \argmin_{\alpha\!\in\!\mathbb{R},\weights\in \mathbb{R}^{\featurelen} }  \sum_{\sampleidx=1}^{\samplesize}  
			\underbrace{\big(\truelabel^{(\sampleidx)} \!-\! \weights^{T} \features^{(\sampleidx)} \big)^{2}}_{\mbox{empirical risk}} \nonumber \\ 
			&\hspace*{20mm}+ \eermparam \underbrace{\big( \weights^{T} \features^{(\sampleidx)} - \alpha \user^{(\sampleidx)} \big)^{2}}_{\mbox{subjective explainability}} 
		\end{align}
		\State {\bfseries Output:} $h^{(\eermparam)}(\features) \defeq \features^{T} \widehat{\mathbf{w}}$ 
	\end{algorithmic}
\end{algorithm}

{\bf Fundamental Trade-Off Between Subjective Explainability and Risk.} 
Let us now study the fundamental trade off between subjective explainability $\subjectiveexplainability{\hypothesis}{\user}$ 
and risk of a linear hypothesis for data points characterized by a single feature $\feature$. 
We consider data points $(\feature,\user,\truelabel)^{T}$, characterized by a single feature $\feature \in \mathbb{R}$, numeric 
label $\truelabel \in \mathbb{R}$ and user feedback $\user \in \mathbb{R}$, as i.i.d.\ realizations of a Gaussian random vector   
\begin{equation} 
\label{equ_prob_model_limit}
\big(\feature,\truelabel,\user)^{T} \sim \mathcal{N}\big( \boldsymbol{\mu}, \mathbf{C} \big) \mbox{ with } \boldsymbol{\mu} = \begin{pmatrix} \mu_{\feature} \\ \mu_{\truelabel} 
\\ \mu_{\user} \end{pmatrix},  
\mathbf{C} = \begin{pmatrix} \sigma_{\feature}^{2} & \sigma_{\feature,\truelabel} & \sigma_{\feature,\user} \\ \sigma_{\truelabel,\feature} & \sigma_{\truelabel}^{2} & \sigma_{\truelabel,\user} 
\\ \sigma_{\user,\feature} & \sigma_{\user,\truelabel} & \sigma^2_{\user} \end{pmatrix}.
\end{equation} 


Our goal is to learn a linear hypothesis $\hypothesis(\features) = \features^{T} \weights$ which is parametrized 
by a weight vector $\weights = \begin{pmatrix} \weight_{1} & \weight_{0} \end{pmatrix}^{T}$, where $\features = \begin{pmatrix} \feature & 1  \end{pmatrix}^{T}$. 
Let us require a minimum prescribed subjective explainability $\subjectiveexplainability{\hypothesis}{\user} \geq C- \eta $, which 
is equivalent to the constraint (see \eqref{eq_def_explainability} and \eqref{equ_sup_mi_Gauss}) 
\begin{align}
	\label{equ_sup_mi_Gauss_eta}
	H(h|\user) &= (1/2) \log  \sigma^{2}_{\hat{\truelabel}|\user} \leq \eta. 
\end{align}
We can further develop the constraint \eqref{equ_sup_mi_Gauss_eta} using \eqref{equ_prob_model_limit} and basic 
calculus for Gaussian processes \cite{Rasmussen2006}, 
\begin{equation}
    \label{cond_cov}
    \begin{aligned}
        \sigma^{2}_{\hat{\truelabel}|\user} &= \sigma^{2}_{\hat{\truelabel}} - \sigma^2_{\hat{\truelabel},\user}/ \sigma^{2}_{\user} \\
        &\stackrel{\hat{\truelabel}= \features^{T} \weights}{=} \weight_{1}^{2}(\sigma^{2}_{\feature} - \sigma^2_{\feature,\user} / \sigma^{2}_{\user}) \\
        &= \weight_{1}^{2}\sigma^{2}_{\feature|\user}.
    \end{aligned}
\end{equation}
The constraint \eqref{equ_sup_mi_Gauss_eta} is enforced by requiring 
\begin{equation}
    \label{equ_constraint_weight_subexp}
    \weight_{1}^{2} \leq \exp(2\eta) \sigma^{-2}_{\feature|\user}. 
\end{equation}
The goal is to find a linear hypothesis $\hypothesis(\feature) =  \features^{T} \weights$, whose weight vector $\weights$ 
satisfies \eqref{equ_constraint_weight_subexp} to ensure sufficient subjective explainability, that incurs minimum risk 
\begin{align} 
\label{equ_def_risk_single_feature}
\risk{\hypothesis} & = \expect \big\{ \big( \truelabel - \hypothesis(\feature) \big)^{2} \big\} \nonumber \\ 
& \stackrel{\hypothesis(\feature)= \features^{T} \weights}{=}   \expect \big\{  (\truelabel - \weight_{1} \feature - \weight_{0})^{2}   \big\}  \nonumber \\
&= \mu_{\feature^2}\weight_1^2 + \weight_0^2 -2\mu_{\truelabel}\weight_0 + \mu_{\truelabel^2} + 2\mu_{\feature}\weight_1\weight_0 - 2\mu_{\feature\truelabel}\weight_1\\
&= \weights^{T} \begin{bmatrix} \mu_{\feature^2} & \mu_{\feature} \\ \mu_{\feature} & 1 \end{bmatrix} \weights - 2\weights^{T} \begin{bmatrix} \mu_{\feature\truelabel} \\ \mu_{\truelabel} \end{bmatrix} + \mu_{\truelabel^2},
\end{align}
where $\mu_{\feature^{2}} = \sigma_{\feature}^{2} + \mu_{\feature}^{2}$ and $\mu_{\feature\truelabel} = \sigma_{\feature,\truelabel} + \mu_{\feature}\mu_{\truelabel}$.

We minimize the risk \eqref{equ_def_risk_single_feature} under the constraint \eqref{equ_constraint_weight_subexp}, 
which is equivalent to enforcing subjective explainability of at least $C - \eta$, 
\begin{align}
\label{equ_opt_risk-constraint}
	\min_{\weight_{1}, \weight_{0} \in \mathbb{R} }& \hspace*{4mm}  \mu_{\feature^2}\weight_1^2 + \weight_0^2 -2\mu_{\truelabel}\weight_0 + \mu_{\truelabel^2} + 2\mu_{\feature}\weight_1\weight_0 - 2\mu_{\feature\truelabel}\weight_1 \\ 
	& \mbox{subject to }   \weight_{1}^{2} \leq \exp(2\eta)\sigma^{-2}_{\feature|\user}
\end{align} 
Any optimal weight $\bar{\weights}$ solving \eqref{equ_opt_risk-constraint} is characterized by the 
Karush-Kuhn-Tucker conditions \cite[Sec. 5.5.3.]{BoydConvexBook}
\begin{align}
\label{kkt_conditions}
 2 \begin{bmatrix} \mu_{\feature^2} + \lagmult & \mu_{\feature} \\ \mu_{\feature} & 1 \end{bmatrix} \weights - 2 \begin{bmatrix} \mu_{\feature\truelabel} \\ \mu_{\truelabel} \end{bmatrix}  & = 0  \nonumber \\ 
\bar{\weight}_{1}^{2} - \exp(2\eta)\sigma^{-2}_{\feature|\user} & \leq 0 \nonumber \\ 
 \lagmult & \geq 0 \nonumber \\ 
\lagmult \big( \bar{\weight}_{1}^{2} - \exp(2\eta)\sigma^{-2}_{\feature|\user}  \big) & = 0. 
\end{align}
The quantity $\lagmult$ appearing in these optimality conditions is known as a Lagrange multiplier \cite[Ch. 5]{BoydConvexBook}. 
By inspection of \eqref{kkt_conditions},  one can show that
\begin{align}
	\label{equ_opt_weight_linear_eerm} 
&\bar{\weight}_{1} = \begin{cases}   \sigma_{\truelabel,\feature}/ \sigma_{\feature}^{2}  & \mbox{ if }   \sigma^2_{\truelabel,\feature}/ \sigma_{\feature}^{4} \leq \exp(2\eta)\sigma^{-2}_{\feature|\user}  \\  {\rm sign} \{ \sigma_{\truelabel,\feature} \} \exp(\eta)/\sigma_{\feature|\user}  & \mbox{ if }  \sigma^2_{\truelabel,\feature}/ \sigma_{\feature}^{4} > \exp(2\eta)\sigma^{-2}_{\feature|\user}. \end{cases}  \\
       \label{equ_opt_weight_const_linear_eerm}
&\bar{\weight}_{0} = \mu_{\truelabel} - \mu_{\feature} \bar{\weight}_{1}
\end{align}

By inserting \eqref{equ_opt_weight_linear_eerm} \eqref{equ_opt_weight_const_linear_eerm} into \eqref{equ_def_risk_single_feature}, we obtain that 
the minimum achievable risk $\risk{\hypothesis}$ of a linear hypothesis with required subjective 
explainability $\subjectiveexplainability{\hypothesis}{\user} \geq C - \eta$ is 
\begin{equation} 
		\label{equ_minimum_risk_subjexpl_linear_eerm} 
	\risk{\hypothesis} = \begin{cases}   \sigma^2_{\truelabel|\feature} & \mbox{ if }   \sigma^2_{\truelabel,\feature}/ \sigma_{\feature}^{4} \leq \exp(2\eta)\sigma^{-2}_{\feature|\user}  \\ 
		
	 \sigma_{\feature}^{2}	 \exp(2\eta)\sigma^{-2}_{\feature|\user} - 2|  \sigma_{\truelabel,\feature}| \exp(\eta)/\sigma_{\feature|\user} +\sigma^2_{\truelabel} & \mbox{ if }  \sigma^2_{\truelabel,\feature}/ \sigma_{\feature}^{4} > \exp(2\eta)\sigma^{-2}_{\feature|\user}. \end{cases}  
\end{equation}

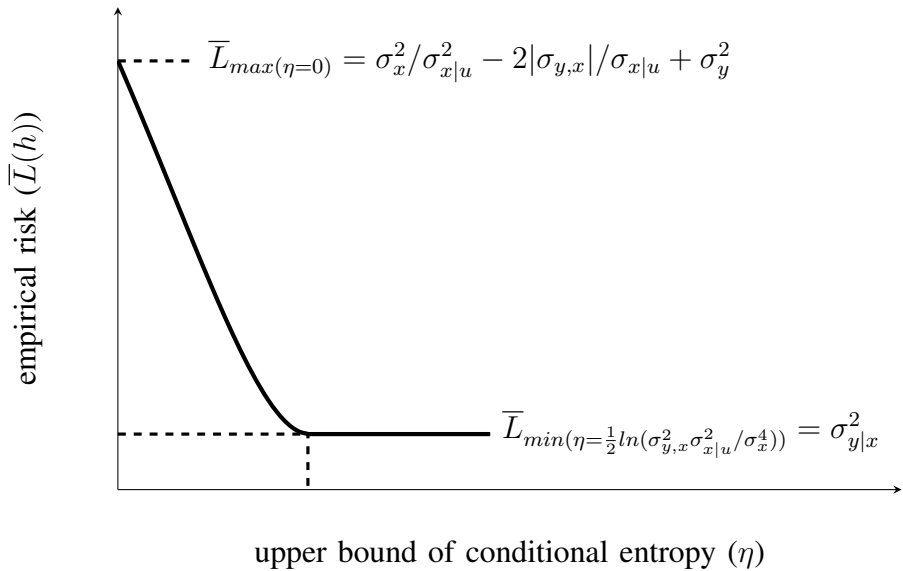
\begin{figure}[htpb]
	\begin{center}

\tikzset{global scale/.style={
    scale=#1,
    every node/.append style={scale=#1}
  }
}
\begin{tikzpicture}[global scale = 1]          
   
        \begin{axis}[width=12cm, height=8cm, at={(0,0)}, ymin=10, ymax=55, xmin=0, xmax=4, ticks=none, ylabel={empirical risk ($\overline{L}(\hypothesis)$)}, xlabel={upper bound of conditional entropy ($\eta$)}, ylabel style={font=\fontsize{12}{12}\selectfont}, xlabel style={font=\fontsize{12}{12}\selectfont}, axis lines = left, axis x line*=middle, axis y line*=none]
        \addplot[color=black, smooth, tension=0.7, ultra thick, domain=0:0.97] {12.77*e^(2*x)-2*33.85*e^(x)+104.93};
        \addplot[color=black, smooth, tension=0.7, ultra thick, domain=0.97:1.9] {15.18};
        \addplot[color=black, very thick, dashed] coordinates{(0,49.99)(0.4,49.99)} node[right, font=\fontsize{12}{12}\selectfont] {$\overline{L}_{max(\eta=0)}=\sigma_{\feature}^{2} /\sigma^{2}_{\feature|\user} - 2|  \sigma_{\truelabel,\feature}| /\sigma_{\feature|\user} +\sigma^2_{\truelabel}$};
        \addplot[color=black, very thick, dashed] coordinates{(0,15.18)(1.9,15.18)} node[right, font=\fontsize{12}{12}\selectfont] {$\overline{L}_{min(\eta=\frac{1}{2}ln(\sigma^2_{\truelabel,\feature} \sigma^{2}_{\feature|\user}/ \sigma_{\feature}^{4}))}=\sigma^2_{\truelabel|\feature}$};
        \addplot[color=black, very thick, dashed] coordinates{(0.97,0)(0.97,15.18)};
    \end{axis}

\end{tikzpicture}
		\vspace*{-3mm}
		\caption{The solutions of EERM, in the dual \eqref{equ_def_eerm_constrained} form, 
			trace out a curve in the plane spanned by the risk $\overline{L}(h)$ and the upper bound of conditional entropy $\eta$.}
		\label{fig:duallinregreerm}
		\vspace*{-1mm}
	\end{center}
\end{figure}

\subsection{Explainable Decision Trees}
\label{sec_dectree}

We now specialize EERM in its dual (constraint) form \eqref{equ_def_eerm_constrained} to decision tree 
classifiers \cite{BishopBook,HastieWainwrightBook}. Consider data points characterized by features $\features$ 
and a binary label $\truelabel \in \{0,1 \}$. Moreover, each data point is characterized by a binary 
user signal $\user \in \{0,1\}$. The restriction to binary labels and user signals is for ease of exposition. 
Our approach can be generalized easily to more than two label values (mult-class classification) and 
non-binary user signals. 

The model $\mathcal{H}$ in \eqref{equ_def_eerm_constrained} is constituted by 
all decision trees whose root node tests the user signal $\user$ and whose depth 
does not exceed a prescribed maximum depth $\treedepth_{\rm max}$ \cite{hastie01statisticallearning}. 
The depth $\treedepth$ of a specific decision tree $h$ is the maximum number of 
test nodes that are encountered along any possible path from root node to a leaf node \cite{hastie01statisticallearning}.
 
\begin{figure}[htpb]
	\begin{center}
		\includegraphics[width=0.9\columnwidth]{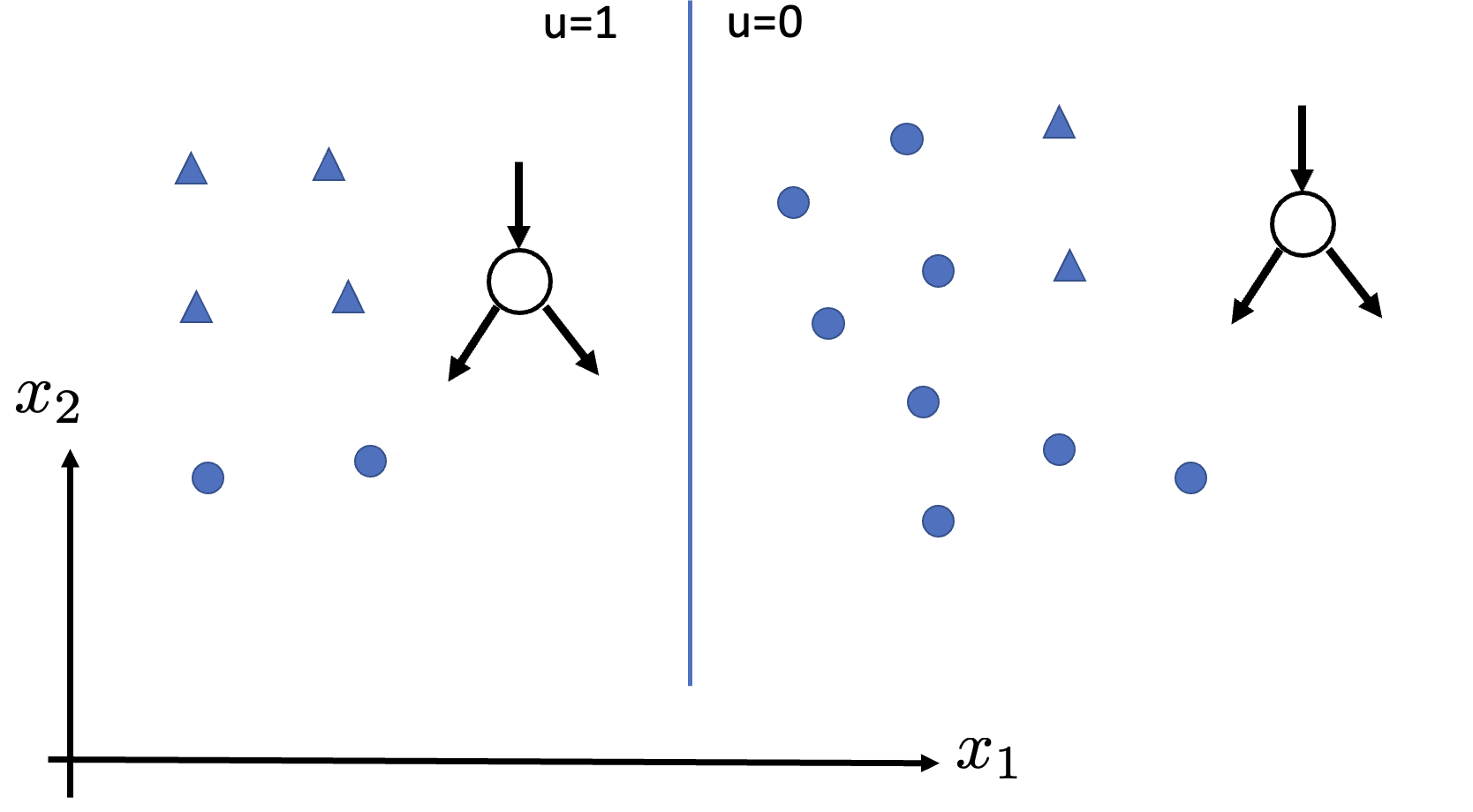}
		\vspace*{-3mm}
		\caption{EERM implementation for learning an explainable decision tree classifier. EERM amounts to learning 
			a separate decision tree for all data points sharing a common user signal $\user$. 
		The constraint in \eqref{equ_def_eerm_constrained} can be enforced naturally by fixing a maximum tree depth $\treedepth$.}
		\label{fig:dteerm}
		\vspace*{-1mm}
	\end{center}
\end{figure}

Figure \eqref{fig:dteerm} illustrates a hypothesis $h$ obtained from a decision tree with depth $\treedepth=2$. 
We consider only decision trees whose nodes implement a binary test, such as whether a 
specific feature $\feature_{\featureidx}$ exceeds some threshold. Each such binary test can 
maximally contribute one bit to the entropy of the resulting prediction (at some leaf node). 

Thus, for a given user signal $\user$, the conditional entropy of the prediction $\hat{\truelabel} = h(\features)$ 
is upper bounded by $\treedepth-1$ bits. Indeed, since the root node is reserved for testing the user signal $\user$, 
the number of binary tests carried out for computing the prediction is upper bounded by $\treedepth-1$. 
We then obtain Algorithm \ref{alg:explainable_dtclass} from \eqref{equ_def_eerm_constrained} by using the estimator 
$\widehat{H}(h|\user)  \defeq \treedepth -1$.

\begin{algorithm}[htbp]
	\caption{Explainable Decision Tree Classification}\label{alg:explainable_dtclass}
	\begin{algorithmic}
		\State {\bfseries Input:} subjective explainability $\eta$, 
		training set $\dataset$ \eqref{equ_def_dataset}
			\vspace*{2mm}
		\State 1: maximum tree-depth $\treedepth_{\rm max} \defeq \lceil \eta \rceil$
			\vspace*{2mm}
		\State 2: partition dataset $\dataset = \dataset^{(\user=0)}  \bigcup \dataset^{(\user=1)}$ with 
		\begin{align}
		 \dataset^{(\user=0)} & \defeq \big\{ \big( \features^{(\sampleidx)}, \truelabel^{(\sampleidx)}, \user^{(\sampleidx)} \big): \user^{(\sampleidx)}=0 \big\} 
		\subseteq \dataset \nonumber \\
		 \dataset^{(\user=1)} & \defeq \big\{ \big( \features^{(\sampleidx)}, \truelabel^{(\sampleidx)}, \user^{(\sampleidx)} \big): \user^{(\sampleidx)}=1 \big\} 
		\subseteq \dataset
		\end{align} 
	\vspace*{2mm}
		\State 3: learn decision tree classifier $h^{(\user=0)}$ with maximum depth $\treedepth_{\rm max}$ using training set $\dataset^{(\user=0)}$
			\vspace*{2mm}
		\State 4 learn decision tree classifier $h^{(\user=1)}$ with maximum depth $\treedepth_{\rm max}$ using training set $\dataset^{(\user=1)}$
			\vspace*{2mm}
		\State {\bfseries Output:} $h^{(\eta)}(\features) \defeq \begin{cases} h^{(\user=1)}(\features) & \mbox{ if } \user = 1 \\ h^{(\user=0)}(\features) & \mbox{ if } \user = 0 \end{cases}$ 
	\end{algorithmic}
\end{algorithm}


\section{Numerical Experiments}
\label{sec_num_exp}

\subsection{Explainable Linear Regression}

Firstly we study the usefulness of EERM by numerical experiments revolving around the problem of predicting the weather temperature based on a public dataset during 01.01.2020 - 13.12.2021 downloaded from Finnish Meteorological Institute (FMI) (https://en.ilmatieteenlaitos.fi/download-observations). 

Specifically, we consider datapoints that represent the daily weather recordings along with a time-stamp at Nuuksio in Finland. The feature vector $\features$ is constructed using the numerical value of the minimum temperature of a day while the maximum temperature of the same day is the label $\truelabel$. Each datapoint is also characterized by a user signal $\user \in \mathbb{R}$. The user signal is defined as the maximum temperature with perturbations.

We employ the formulas \eqref{equ_opt_weight_linear_eerm} - \eqref{equ_minimum_risk_subjexpl_linear_eerm} to learn an explainable linear regression with its subjective explainability upper bounded via a given value of $\eta$. The Gaussian random vector is calculated by datapoints in the training set.

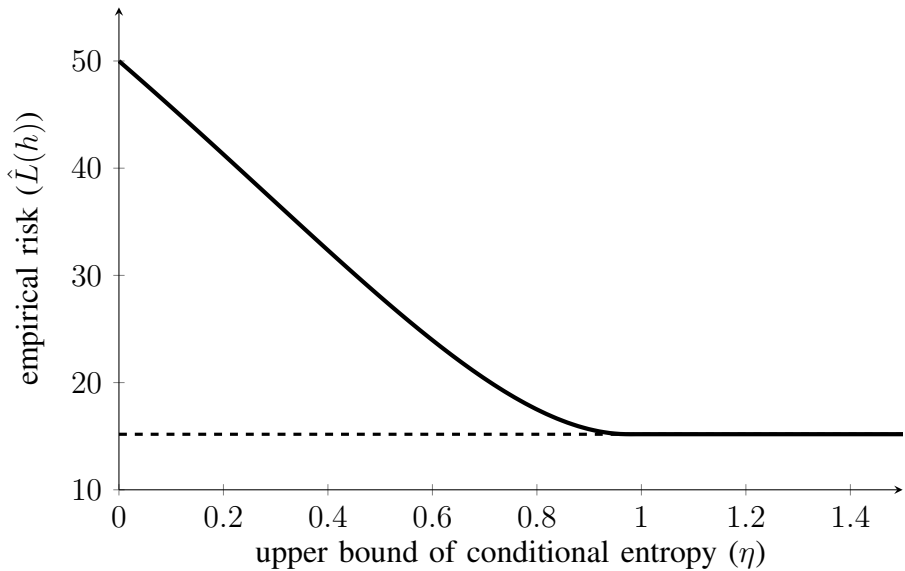
\begin{figure}[h]
	\begin{center}
		\tikzset{global scale/.style={
    scale=#1,
    every node/.append style={scale=#1}
  }
}
		\begin{tikzpicture}[global scale = 1]          
    
                     \begin{axis}[width=12cm, height=8cm, at={(0,0)}, ymin=10, ymax=55, xmin=0, xmax=1.5, ylabel={empirical risk ($\hat{L}(\hypothesis)$)}, xlabel={upper bound of conditional entropy ($\eta$)}, ylabel style={font=\fontsize{12}{12}\selectfont}, xlabel style={font=\fontsize{12}{12}\selectfont}, axis lines = left, axis x line*=middle, axis y line*=none]
                         \addplot[color=black, smooth, tension=0.7, ultra thick] table [x =eta, y =l,col sep=comma] {lrpararesults.csv};
                         \addplot[color=black, very thick, dashed] coordinates{(0,15.18)(2,15.18)};
                     \end{axis}
    
\end{tikzpicture}
		
		\vspace*{-3mm}
		\caption{the subjective explainability $E(h|\user)$ influences the empirical risk $\hat{L}(h)$ via a given value of $\eta$.}
		\label{fig:eermLRrisk}
		\vspace*{-1mm}
	\end{center}
\end{figure}

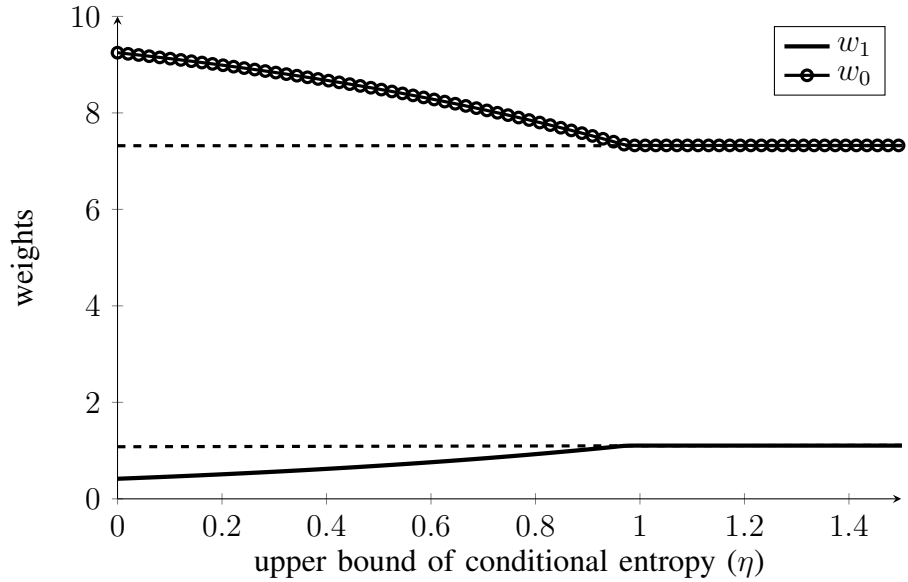
\begin{figure}[h]
	\begin{center}
		
		\tikzset{global scale/.style={
    scale=#1,
    every node/.append style={scale=#1}
  }
}
\begin{tikzpicture}[global scale = 1] 
    
    \begin{axis}[width=12cm, height=8cm, at={(0,0)}, ymin=0, ymax=10, xmin=0, xmax=1.5, ylabel={weights}, xlabel={upper bound of conditional entropy ($\eta$)}, ylabel style={font=\fontsize{12}{12}\selectfont}, xlabel style={font=\fontsize{12}{12}\selectfont}, axis lines = left, axis x line*=middle, axis y line*=none]
        \addplot[color=black, smooth, tension=0.7, ultra thick] table [x =eta, y =w, col sep=comma] {lrpararesults.csv};
        \addplot[color=black, mark=o, smooth, tension=0.7, very thick] table [x =eta, y =b, col sep=comma] {lrpararesults.csv};
        \addplot[color=black, very thick, dashed] coordinates{(0,1.08)(1.5,1.11)};
        \addplot[color=black, very thick, dashed] coordinates{(0,7.32)(1.5,7.32)};
        
        \addlegendentry{$w_1$};
        \addlegendentry{$w_0$};
    \end{axis}
    
\end{tikzpicture}
		
		\vspace*{-3mm}
		\caption{the subjective explainability $E(h|\user)$ influences the optimal weights $\weights$ via a given value of $\eta$.}
		\label{fig:eermLRweight}
		\vspace*{-1mm}
	\end{center}
\end{figure}

The results in Figure \eqref{fig:eermLRrisk} and \eqref{fig:eermLRweight} show that as the upper bound of the conditional entropy $\eta$ increases, i.e., the subjective explainability $E(h|\user)$ decreases, the empirical risk $\hat{L}(h)$ goes down until the optimal weight $\weights$ is achieved. 

\subsection{Explainable Decision Tree}
Then we study the usefulness of EERM by numerical experiments revolving around the 
problem of detecting hate-speech and offensive language in social media \cite{WangTwitter2011}. 
Hate-speech is a contested term whose meaning ranges from concrete threats to individuals to 
venting anger against authority \cite{Gagliardone2015}. Hate-speech is characterized 
by devaluing individuals based on group-defining characteristics such as their race, 
ethnicity, religion and sexual orientation \cite{Erjavec2012}. 

Our experiments use a public dataset that contains curated short messages (tweets) from a 
social network \cite{Davidson2017}. Each tweet has been manually rated by a varying number of 
users as either ``hate-speech'', ``offensive language'' or ``neither''. For each tweet we define 
its binary label as $\truelabel=1$ (``inappropriate tweet''') if the majority of users rated the tweet 
either as ``hate-speech'' or ``offensive language''. If the majority of users rated the tweet 
as ``neither'', we define its label value as $\truelabel=0$  (``appropriate tweet''). 

The feature vector $\vx$ of a tweet is constructed using the normalized frequencies (``tf-idf'') 
of individual words \cite{Yates2011}. Each tweet is also characterized by a binary user signal $\user \in \{0,1\}$. 
The user signal is defined to be $\user=1$ if the tweet contains at least one of the $5$ most 
frequent words appearing in tweets with $\truelabel=1$. 

We use Algorithm \ref{alg:explainable_dtclass} to learn an explainable decision tree classifier 
with its subjective explainability upper bounded by $\eta=2$ bits. The training set $\dataset$ 
used for Algorithm \ref{alg:explainable_dtclass} is obtained by randomly selecting a fraction 
of around $90  \%$ percent of the entire dataset. The remaining $10 \%$ of tweets are used 
as a test set. 

To learn the decision tree classifiers in step $3$ and $4$ of Algorithm \ref{alg:explainable_dtclass}, 
we used the implementations provided by the current version of the Python package \texttt{scikit-learn} \cite{JMLR:v12:pedregosa11a}. 
The resulting explainable decision tree classifier (with the root node testing the user signal) 
$h^{(\eta=2)}(\features)$ achieved an accuracy of $0.929$ on the test set.

\section{Conclusion} 
The explainability of predictions provided by ML becomes increasingly relevant for their 
use in automated decision-making \cite{Rohlfing2021,Larsson2020}. 
Given lay and expert user’s different level of expertise and knowledge, providing subjective (tailored) 
explainability is instrumental for achieving trustworthy AI \cite{Sokol2020,Rohlfing2021}.
Our main contribution is EERM as a new design principle for subjective explainable ML. EERM is obtained 
by using  the conditional entropy of predictions, given a user signal, as a regularizer. 
The hypothesis learned by EERM balances between small risk and a sufficient explainability 
for a specific user (explainee).  



\bibliographystyle{IEEEtran}
\bibliography{Literature}

\newpage
\onecolumn
%

\end{document}